\documentclass[10pt,twocolumn,letterpaper]{article}

\usepackage[e.g.]{cvpr}      % To produce the REVIEW version
\definecolor{cvprblue}{rgb}{0.21,0.49,0.74}
\usepackage[pagebackref,breaklinks,colorlinks,allcolors=cvprblue]{hyperref}
\usepackage{makecell}
\usepackage{multirow}
\usepackage{wasysym}
\usepackage[table]{xcolor} 

\def\paperID{*****} % *** Enter the Paper ID here
\def\confName{CVPR}
\def\confYear{2026}

\title{Beyond Imitation: Constraint-Aware Trajectory Generation with Flow Matching For End-to-End Autonomous Driving}%songziying

\author{Lin Liu$^{1,2}$, Guanyi Yu$^{2}$, Ziying Song $^{1}$, JunQiao Li$^{2}$, Caiyan Jia$^{1}$, Feiyang Jia$^{1}$, Peiliang Wu$^{3}$, Yandan Luo$^{4}$\vspace{1ex}\\
$^1$Beijing Jiaotong University
$^2$Qcraft
$^3$Yanshan University
$^4$The University of Queensland
\\
\tt\small\{23120379, songziying, cyjia\}@bjtu.edu.cn.
\vspace{-1ex}
}

\begin{document}
\maketitle
\begin{abstract}
Planning is a critical component of end-to-end autonomous driving. However, prevailing imitation learning methods often suffer from mode collapse, failing to produce diverse trajectory hypotheses. Meanwhile, existing generative approaches struggle to incorporate crucial safety and physical constraints directly into the generative process, necessitating an additional optimization stage to refine their outputs. To address these limitations, we propose CATG, a novel planning framework that leverages Constrained Flow Matching. Concretely, CATG explicitly models the flow matching process, which inherently mitigates mode collapse and allows for flexible guidance from various conditioning signals. Our primary contribution is the novel imposition of explicit constraints directly within the flow matching process, ensuring that the generated trajectories adhere to vital safety and kinematic rules. Secondly, CATG parameterizes driving aggressiveness as a control signal during generation, enabling precise manipulation of trajectory style. Notably, on the NavSim v2 challenge, CATG achieved \textbf{2nd} place with an EPDMS score of 51.31 and was honored with the \textbf{Innovation Award}.
%Trajectory generation is crucial for end-to-end autonomous driving. However, as one of the mainstream paradigms, imitation learning suffers from mode collapse and struggles to generate diverse multi-modal trajectories. Another paradigm for trajectory prediction relies on generative models but typically necessitates a separate refinement stage to adjust the outputs. This necessity arises from the inherent difficulty of incorporating crucial safety and physical constraints during the generation process. To address these issues, this work proposes a novel trajectory generation framework named CATG (Constraint-Aware Trajectory Generation), which leverages the capabilities of flow matching for trajectory generation and completely eliminates the imitation learning-based refinement stage. CATG incorporates the following innovations: (1) It leverages a flow matching framework for trajectory generation, supports control through multiple conditional variables, and completely eliminates imitation learning from its training process. (2) It seamlessly integrates hard constraints into the generation process, ensuring that trajectories satisfy key safety and physical requirements. (3) It treats environment-provided trajectory rewards as conditional inputs, enabling flexible switching between aggressive and conservative behaviors during inference. By integrating our trajectory generation model with existing open-source scoring models, our approach achieves state-of-the-art performance on the NAVSIM V2 end-to-end driving benchmark, with a score of 51.31 EPDMS.
\end{abstract}
% 规划是端到端自动驾驶最为关键的环节之一。然而，主流模仿学习方法常因模式坍缩而难以输出多模态轨迹；而现有基于生成模型的方法则因在生成过程中难以有效融入安全与物理约束,从而需要额外的优化阶段来调整输出。为了解决上述问题，本文提出CATG，which 利用Constrained-Flow-Matching进行规划。具体的, ConstrainedDrive显式建模流匹配过程，从而避免了模仿学习所面临的模式坍缩问题并可接受多种signal进行生成控制。ConstrainedDrive首次创新地尝试直接显式约束流匹配过程从而使得轨迹满足关键的安全和物理约束。其次，ConstrainedDrive将轨迹的激进程度建模为生成过程中的控制信号，从而实现轨迹风格的控制。此外，ConstrainedDrive尝试减少采样步长进行加速推理从而提高方法的实用性。ConstrainedDrive在多个驾驶数据集上表现出色，例如bench2drive,nuscenes and navsim数据集。特别地，在navsimv2挑战赛上，CATG取得第二名 with 51.31 EPDMS，并获得innovation reward.
% 轨迹生成是端到端自动驾驶的关键环节。然而，作为主流范式之一，模仿学习存在模式坍缩问题，难以生成多样化多模态轨迹。另一种轨迹预测范式依赖生成式模型，但通常需要额外的优化阶段来调整输出。这种必要性源于在生成过程中融入关键安全与物理约束的固有难度。的新型轨迹生成框架，该框架利用流匹配的能力进行轨迹生成，完全消除了基于模仿学习的细化阶段。CATG融合了以下创新：（1）它利用流匹配框架生成轨迹，支持通过多个条件变量进行控制，并完全消除了训练过程中的模仿学习。（2）它将硬约束无缝集成到生成过程中，确保轨迹满足关键的安全和物理要求。（3）它将环境提供的轨迹奖励视为条件输入，在推理过程中能够在攻击性和保守性行为之间灵活切换

\section{Introduction}
\label{sec:intro}

% 端到端多模态规划已成为自动驾驶领域的一种强大方法。与预测单个轨迹相比，多模态方法生成多个候选轨迹，从而在推理过程中具有更大的鲁棒性以及适应性,尤其是在可能存在多个轨迹最优解的场景中。 然而，先前的方法多依赖于使用模仿学习进行多模态轨迹预测，由于被学习GT轨迹的单一特性，模型学习出的多模态轨迹通常具有高度的相似性，并极难在推理阶段动态调整。另一类方法尝试引入生成模型的范式进行轨迹预测，然而并不显式监督加噪过程，这类方法尽管收益于生成模型但是由于仍然使用模仿学习同样面临着多模态轨迹模式坍缩的问题。最后一类方法，完全遵循生成模型的范式进行训练以及推理，然而由于生成起点采样的随机性以及生成过程难以进行显式约束的问题, 生成模型同样面临着安全以及驾驶合规性的挑战。
% 为了解决上述问题，我们提出了CATG，其使用更为灵活flow matching作为轨迹生成框架，并在训练过程中完全去除了模仿学习，同时为生成过程灵活地引入了显示约束，具体而言我们的CATG包含三个核心创新：（1）CATG依赖flow matching作为轨迹生成框架，在去除模仿学习的同时，灵活地支持多种条件控制多模态轨迹生成，从而避免模式坍缩。（2）CATG从软到硬地将显式约束结合到生成过程中，从使用先验anchor设计一条约束引导的概率流到利用能量模型引导轨迹朝着符合约束的区域前进。 (3) CATG将环境给与轨迹的奖励视为条件信号，从而支持在推理过程中驾驶风格在激进和保守的转变。CATG具体的贡献如下：我们提出了CATG一种可以接收多种灵活条件作为控制信号的多模态轨迹生成器，并去除了传统方法中的模仿学习，此外，我们将显式约束引入到生成过程中，从而使轨迹满足特定的约束。
% CATG在ICCV Navsim v2挑战赛上展示了对域外数据更好的规划性能和泛化。通过和已有的打分器模型GTRS集成，我们的方法接近最先进方法的性能，并获得了51.13的EPDMS分数

End-to-end multimodal planning~\cite{sun2024sparsedrive,liao2024diffusiondrive,jiang2023vad,chen2024vadv2} has established itself as a critical methodology in autonomous driving systems, significantly enhancing robustness and adaptability during inference when compared to single-trajectory prediction approaches. This capability is especially vital in ambiguous or highly interactive driving scenarios—such as unprotected left turns, merging in dense traffic, or navigating intersections—where multiple distinct trajectories may be equally appropriate. Despite these advantages, the majority of contemporary multimodal methods remain dependent on imitation learning frameworks. Such approaches~\cite{jiang2023vad,chen2024vadv2,chen2025ppad,momad,sun2024sparsedrive,uniad,thiktwice} learn from a limited set of demonstrated expert trajectories, and due to the lack of strategy diversity of ground-truth trajectories, often yield predictions that are homogenized, and deficient in behavioral diversity. 

In response to these shortcomings, several alternative strategies have been proposed. A series of works incorporates generative models, such as diffusion processes, to capture a broader distribution of plausible trajectories. However, many of these methods~\cite{liao2024diffusiondrive,goalflow} do not explicitly supervise the generative denoising process, still relying heavily on behavior cloning objectives. As a result, they remain susceptible to mode collapse. Another paradigm~\cite{HE-Drive,zheng2024genad,diffusionplanner} represents a further shift, depending entirely on generative models for trajectory planning and abandoning the use of imitation learning. While these methods benefit from generative models, they introduce new challenges: the stochasticity in noise initialization can lead to high-variance predictions, and the absence of a mechanism for hard constraint integration, such as obstacle avoidance or compliance with traffic rules, compromises the safety and interpretability of generated trajectories. 

To address these limitations, we propose CATG, a novel trajectory generation framework based on flow matching that completely eliminates imitation learning while enabling flexible injection of explicit constraints into the generative process. Our contributions are threefold: 

(1) \textbf{Novel generative framework.} We introduce CATG, a multimodal trajectory generator built upon flow matching. Unlike conventional methods, CATG eliminates the reliance on imitation learning while supporting diverse and flexible conditional controls.

(2) \textbf{Constraint-guided generation.} We explicitly integrate feasibility and safety constraints into the generative process through a progressive mechanism: prior-informed anchor design is used to construct constraint-guided probability flows, and energy-based guidance further steers trajectories toward feasible regions.

(3) \textbf{Reward-conditioned controllability.} We treat environmental reward signals as conditional inputs, enabling controllable trade-offs between aggressive and conservative driving styles during inference.

CATG is extensively evaluated on the ICCV NAVSIM V2 End-to-End Driving Challenge, where it demonstrates superior planning accuracy and robust generalization to out-of-distribution data. When combined with an open-source scoring model, CATG achieves an EPDMS score of 51.31, competitive with state-of-the-art alternatives.

%(1) CATG leverages flow matching as its foundational generative framework, supporting diverse conditional controls for multimodal trajectory generation without imitation learning. (2) CATG incorporates explicit constraints in a progressive manner, which using prior-informed anchor design to construct constraint-guided probability flows and energy-based guidance to steer trajectories toward feasible and safe regions. (3) CATG treats environmental reward signals as conditional inputs, enabling controllable trade-offs between aggressive and conservative driving styles during inference.

%Overall, the primary contributions of CATG are as follows: 

%(1) We introduce CATG, a novel multimodal trajectory generator capable of incorporating diverse and flexible conditions. This framework fundamentally departs from conventional approaches by eliminating the reliance on imitation learning. 

%(2) We explicitly integrate hard constraints directly into the generative process, ensuring that the generated trajectories adhere to specific feasibility and safety requirements. The efficacy of CATG is demonstrated on the ICCV NAVSIM V2 benchmark, where it exhibits superior planning performance and enhanced generalization capabilities on out-of-distribution data. When integrated with the established open-source scoring model, our method achieves performance competitive with state-of-the-art alternatives, attaining an EPDMS score of 51.31.

\section{Preliminary}
\label{sec:pre}
Let $\mathbb{R}^{d}$ denote the data space, two important objects we use in this paper are: the probability density path $p$ : $[0, 1] \times \mathbb{R}^{d} \to \mathbb{R}_{>0}$, which is a time dependent probability density function i.e., $\int p_{t}(x)dx = 1$, and a time-dependent vector field, $v:[0,1]\times\mathbb{R}^{d} \to \mathbb{R}^{d}$. A vector field $v_{t}$ can be used to construct a time-dependent diffeomorphic map, called a flow, $\phi:[0,1]\times\mathbb{R}^{d}\to\mathbb{R}^{d}$. This flow serves as a probability path $p_{t}(x)$ connecting the source distribution $X_{0} \sim \pi_{0}$ and target distribution $X_{1} \sim \pi_{1}$, defined via the ordinary differential equation (ODE):
\begin{align}
    \frac{d}{dt}\phi_{t} (x) &= v_{t}(\phi_{t}(x))\\
    \phi_{0} (x) &= x
\end{align}

And, we can model the vector field $v_{t}$ with a neural network, $v_{t}(t;\theta)$. Let $X_{1}$ denote a random variable distributed according to an unknown data distribution $\pi_{1}$. We assume that we only have access to data samples from $\pi_{1}$, but not to the density function itself. Furthermore, we let $\pi_{0}$ be a simple distribution, such as a standard normal distribution. Given a target probability density path $p_{t}(x)$ and a corresponding vector field $u_{t}(x)$, which generates $p_{t}(x)$, we define the Flow Matching (FM) objective as:
\begin{align}
    L_{FM}(\theta) = \mathbb{E}_{t,p_{t}(x)}||v_{t}(x) - u_{t}(x)||^{2}
\end{align}
% 令 x1 表示根据某个未知数据分布 q(x1) 分布的随机变量。我们假设我们只能从 q(x1) 访问数据样本，但无法访问密度函数本身。此外，我们让 pt 是一个概率路径，使得 p0 = p 是一个简单的分布，例如标准正态分布
In CATG, we use rectified flow to construct a probability path $\phi$:
\begin{align}
    X_{t} = tX_{1} + (1-t)X_{0}
\end{align}
So, the drift force $v:\pi_{0} \to \pi_{1}$ is set to drive the flow to follow the direction $(X_{1} - X_{0})$ of the linear path pointing from $X_{0}$ to $X_{1}$ as much as possible, by solving a simple least squares regression problem:
\begin{align}
    min_{v}\int_{0}^{1}\mathbb{E}[||(X_{1} - X_{0}) - v(X_{t},t)||^{2}]dt, 
\end{align}
where $X_{t}$ is the linear interpolation of $X_{0}$ and $X_{1}$.

\section{Method}
\label{sec:method}

\begin{figure*}[t]
    \centering
    \includegraphics[width=1.0\linewidth]{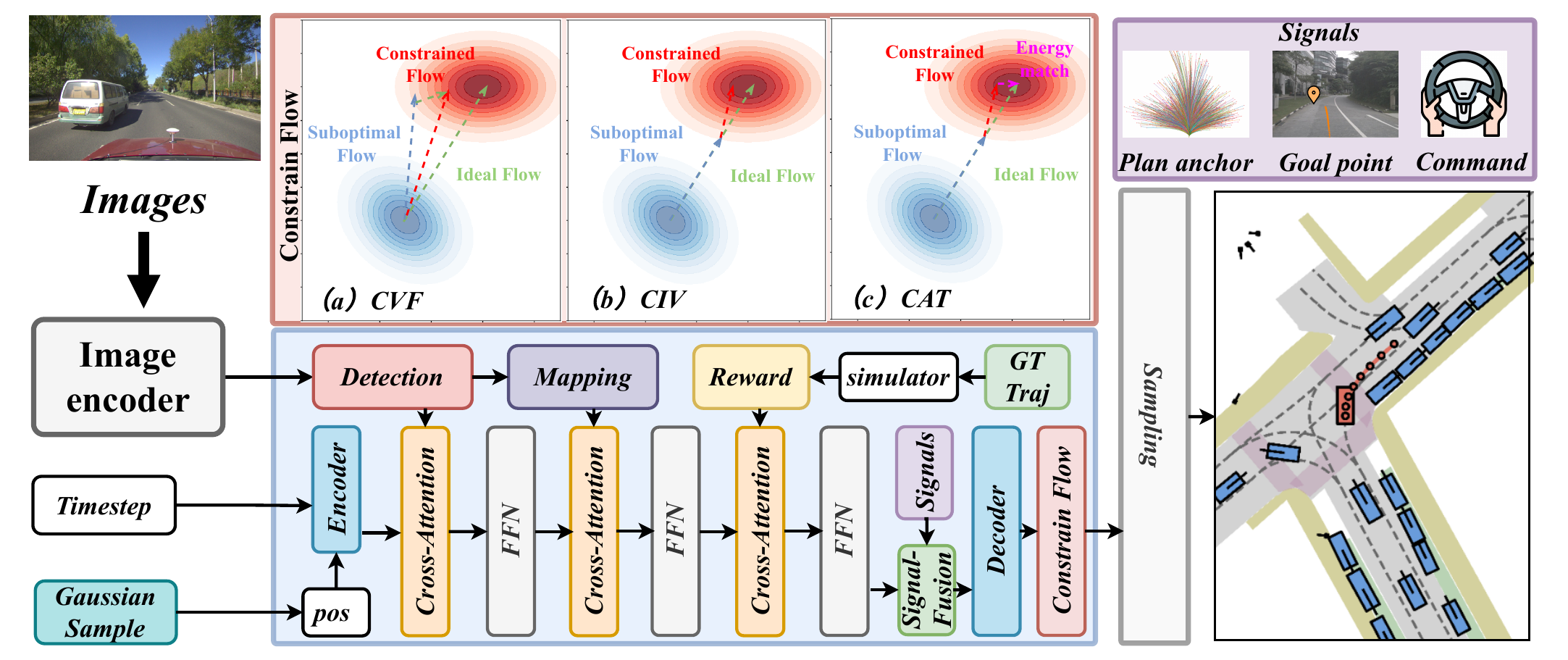}
    \caption{The architecture of CATG with Flow Matching. The image is encoded into image features, which are subsequently fed into detection and mapping modules to generate perceptual features. Before training, CATG processes each gt trajectory (GT Traj) through simulator to compute and offline store its score. Subsequently, CATG encodes Gaussian-sampled latent variables and the timestamp $t$ into input features via a encoder. These features are fused with perceptual features, trajectory rewards, and conditioning signals (including plan anchor, goal point, and driving command) through cross-attention. The signal fusion module consists of multi-layer cross-attention blocks and adheres to the classifier-free~\cite{classifierfreediffusionguidance} training paradigm. The velocity field decoded by the decoder is refined via our three correction strategies: CVF, CIV, and CAT (Sec~\ref{sec:CVF}). Finally, the driving trajectory is generated through sampling.} %Image被编码为高维图像特征随后被送入检测module以及mapping module以产生高维感知特征。在训练前，CATG将每条gt 轨迹送入仿真器中以获得分数并进行离线保存。随后， catg 将高斯采样的sample以及时间戳 t 经过unet-encoder编码为高维特征，并与高维感知特征, 轨迹奖励以及signals (包括 plan anchor, goal point 以及 driving command)}进行cross attention 融合，其中 signal fusion module 由多层的cross attention组成，并遵循classifier-free 的训练范式。decoder解码出的速度场将被我们的三类纠正方式改进，分别为CVF, civ以及CAT.最终，通过采样得到驾驶轨迹
    \label{fig:main}
\end{figure*}

\subsection{Flexible conditioning signal}
We followed the Transfuser~\cite{TransFuser} as our perception backbone. For flow matching progress, we sample $X_{0}$ from a standard Gaussian distribution and normalize the target trajectory $X_{1}$ to the range $[-1, +1]$. CATG constructs a flow with the starting point as $X_{0}$ and the endpoint as$X_{1}$. Then, we apply positional encoding to $X_{t}$ and utilize a Unet Encoder~\cite{MCG_diffusion} to encode $X_{t}$ into a feature $F_{X_{t}}$. Subsequent to the CATG perception module, CATG obtains the agent's query $Q_{ag}$, ego query $Q_{eg}$, and BEV feature $F_{B}$. In a separate preprocessing step, the BEV map segmentation result is first converted into a binary road map $M_{0,1}$ and then fused with BEV grid positional encoding $Pos_{B}$. Finally, CATG fuses the feature $F_{X_{t}}$ with all these elements ($Q_{ag}$, $Q_{eg}$, $F_{B}$ and $M_{0,1}$) through multiple layers of cross-attention as shown in Fig.~\ref{fig:com}.
\begin{align}
    F_{X_{t}} &= F_{X_{t}} + Tim_{ebed}(t) \\
    F_{X_{t}} &= CrossAttn(F_{X_{t}}, Q_{ag}) \\
    F_{X_{t}} &= CrossAttn(F_{X_{t}}, Fusion(F_{B},M_{0,1},Pos_{B})) \\
    F_{X_{t}} &= CrossAttn(F_{X_{t}}, Q_{eg})
\end{align}
% 我们从标准高斯分布中采样X_0, 并将目标轨迹X_1进行归一化至[-1,+1]之间，CATG根据flow的起点X_0和终点X_1构建 xt,随后我们将x_{t}进行位置编码并使用Unet Encoder将 X_t 编码为特征 F. After transfuser preception module, CATG gained 智能体的query, ego query and Bev feature. 除此之外，我们将BEV地图分割结果转换为道路二值图，并为其添加BEV 网格位置编码， CATG使用多层cross attention 融合 F 和 智能体的query, ego query ， Bev feature and 道路二值图

% 为了在推理过程中使用classifier-free方式灵活控制轨迹生成样式，我们引入了三种不同类型的条件控制信号:
%(1) traj anchor (2) target point (3) drive command
In order to flexibly control the trajectory generation style in a classifier-free manner~\cite{classifierfreediffusionguidance} during inference, we introduce three distinct types of conditional control signals:

(1) \textbf{Trajectory anchor:}
% 我们将提前聚类得到的anchor视为驾驶模式的高度总结，CATG遍历整个训练集并使用最远点采样的方式采集轨迹词汇表，数量为8192，我们采用classifier-free的训练方式，引入驾驶anchor作为控制轨迹生成的条件，during training, 我们使用dtw距离匹配得到和gt 轨迹最为相似的anchor,作为condition signal, during testing, 我们使用预训练好的打分器GTRS (V2-99 backbone),选择概率最高的100条anchor被用于作为控制信号
CATG treat pre-clustered trajectory anchors as high-level abstractions of driving modes. CATG first constructs a trajectory vocabulary $vocab_{anchor}$ of size 8,192 by applying FPS (farthest-point sampling) over the entire training dataset. CATG is trained in a classifier-free guidance~\cite{classifierfreediffusionguidance} manner, where driving anchors are incorporated as conditional signals to guide trajectory generation. During training, the anchor most similar to the GT trajectory is utilized as the conditional signal, which is determined by DTW distance between trajectory vocabulary and GT trajectory. At inference time, a pre-trained scoring model, GTRS~\cite{GTRS} (with a V2-99 backbone), is employed to select the top-100 anchors with the highest likelihood, which subsequently serve as conditional inputs for generating diverse and compliant trajectories.

(2) \textbf{Target point:}
% during traing, CATG 将gt 轨迹的末端点作为条件信号。 而再推理阶段，我们将打分器获得的anchor 末端点视为条件控制信号
During training, CATG takes the endpoint of the GT trajectory as the conditional signal. During testing, in contrast, the endpoint of the anchor obtained from a scoring model serves as the conditional control signal. 

(3) \textbf{Driving command:}
% 驾驶命令同样作为控制信号的一种，CATG将navsim中的四种驾驶命令转化为one hot编码作为condition signal 使用
The driving command is also a type of control signal. CATG converts the command types in NAVSIM~\cite{dauner2024navsim} into a one-hot encoding for use as a conditional signal.

\subsection{Constraint-Aware Trajectory Generation}
% 生成模型真正困难的是生成的中间结果不具备可解释性，从而难以直接进行约束生成结果，在navsim 挑战中，我们发现很难使得生成轨迹满足DAC指标的约束，这不同于其他约束，如和其他智能体的碰撞，后者可以使用和其他车辆的距离作为条件信号并注入生成过程，就像diffusion planner那样。而道路形状更为复杂，我们使用了两种更为直接且高效的约束方法进行生成约束。 flow matching的生成过程为可以被表述为：
A significant challenge in generative models is the lack of interpretability in their intermediate representations, posing difficulties for directly constraining the outputs. Specifically in the NAVSIM V2~\cite{dauner2024navsim} challenge, constraining the generated trajectories to satisfy the Driving Area Compliance (DAC) metric proved highly challenging. Unlike constraints such as inter-agent collision avoidance which can be integrated by using vehicle distances as conditional signals, as seen in Diffusion-Planner~\cite{diffusionplanner}, road geometry is far more complex. Therefore, in the following discussion, we will primarily focus on constraining trajectories to satisfy road compliance. However, it is noteworthy that our method can also be adapted to other types of constraints. To address this, we introduced three more direct and efficient methods for constraining the generation. The Flow Matching generation process is formulated as:

\begin{align}
    X_{t+1} = X_{t} + v_{t}dt
\end{align}

Since the formulation above indicates that the generated state $X_{t+1}$ at the next timestep is determined by the intermediate variable $X_{t}$ and the velocity field $v_{t}$, a compelling hypothesis arises: could one constrain the generation process by imposing constraints on these two quantities ?
% 由上述公式可知，下一时刻的生成结果由中间过程量X_{t}和速度场V_{t}决定，那是否可以尝试约束两者从而约束生成过程？

(1) \textbf{Constraining velocity field $v_{t}$ (CVF):} \label{sec:CVF} Based on the road segmentation result, a trajectory $X_{1}^{C}$ that satisfying the DAC constraint is first selected from trajectory vocabulary $vocab_{anchor}$. Subsequently, for a given Gaussian sample $X_{0}$ as the flow's starting point, the ideal velocity field that leads to trajectory $X_{1}^{C}$ can be computed. 
\begin{align}
    v_{t}^{c} = \frac{X_{1}^{C} - X_{0}}{1-0}
\end{align}
CATG leverages this precomputed field $v_{t}^{c}$ to correct the potentially biased velocity field $v_{t}$ predicted by the model. Consequently, we propose the concept of a synthetic velocity field $v_{t}^{'}$, which is a combination of the model predicted velocity field $v_{t}$ and the precomputed one $v_{t}^{c}$ during the sampling process as shown in Fig.~\ref{fig:main} (a):

\begin{align}
    v_{t}^{'} = v_{t} +  \frac{2\lambda v_{t} \cdot v_{t}^{c}}{||v_{t}^{c}||^{2}}v_{t}^{c},
\end{align}
where $\lambda$ was set to -0.1.
% (1) v_{t} 约束: 给定一个场景的道路分割结果，我们可以基于此在anchor中选择符合驾驶区域合规性的轨迹 T， 给定生成起始点的高斯采样，我们可以手动计算出该flow的速度场，即，从该高斯采样开始，按此速度场进行采样可获得符合DAC的轨迹。故CATG使用该速度场纠正存在偏差的模型预测出的速度场，因此，我们提出合成速度场的概念，采样过程中的速度场由模型预测，和预先计算的速度场共同得到
% v_{t}^{Synthetic} = v_{t} + lamda v_{t} 其中 lamada被设置为

\begin{figure}[t]
    \centering
    \includegraphics[width=1.0\linewidth]{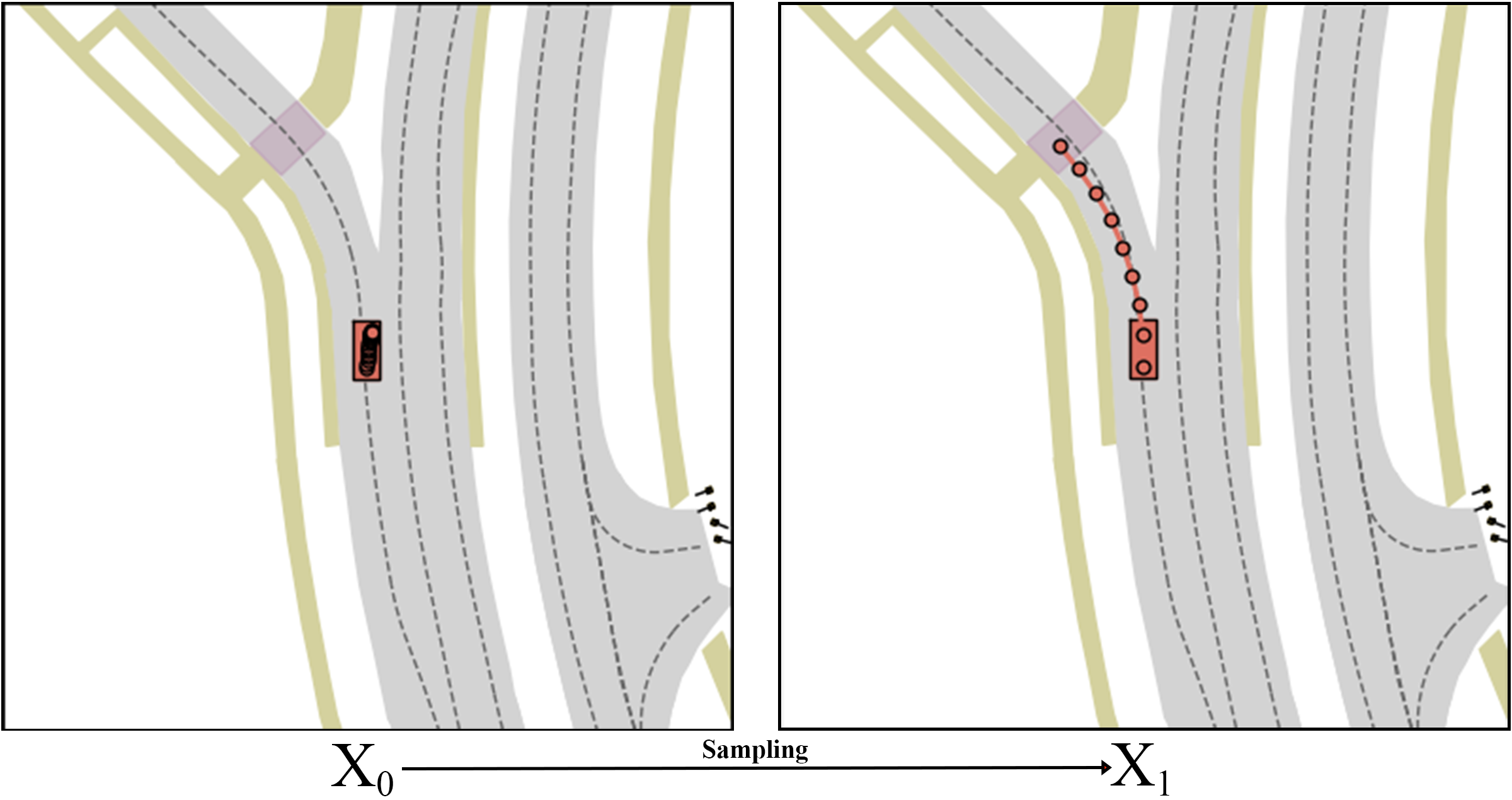}
    \caption{The figure illustrates that sampling by using an anchor as the starting point results in a more reasonable trajectory.}
    \label{fig:com}
\end{figure}

(2) \textbf{Constraining intermediate variables $X_{t}$ (CIV):} A flow generated by a model-predicted velocity field often deviates from the ideal, leading to a final sample that fails to meet constraints. This flow can be discretized into a series of intermediate variables $X_{0},...,X_{t},...,X_{1}$; Therefore, if these intermediate variables can be effectively constrained, the final generated outcome can consequently be controlled. However, correcting $X_{t}$ at every timestep is inefficient. Instead, inspired by~\cite{improvediff}, CATG addresses this by correcting the flow at its origin. It replaces the initial Gaussian random sample $X_{0}$ with an anchor $X_{1}^{C}$ selected from the trajectory vocabulary $vocab_{anchor}$ as shown in Fig.~\ref{fig:main} (b), which complies with the DAC constraint, even though this anchor might perform poorly on other evaluation metrics. However, CATG can refines this anchor to make it more reasonable. As shown in Fig.~\ref{fig:com}, this approach of starting from a DAC-compliant anchor enables the model to produce more plausible trajectories.

(3) \textbf{Constraint-Aware Training (CAT): }In contrast to Diffusion-Planner~\cite{diffusionplanner}, which only introduces energy term during inference, we incorporate constraints into the training phase by encoding them as an energy function. When trajectory are sampled along the direction of ascending energy, they exhibit a higher probability of satisfying the constraints as shown in Fig.~\ref{fig:main} (c). Specifically, the DAC constraint can be represented by computing a Euclidean Signed Distance Field. The energy of a trajectory decreases as it moves closer to the road boundary, penalizing undesirable deviations. We follow the Energy Matching~\cite{energymatching} framework for model training. A two-stage procedure is employed, the first stage trains the Flow Matching process, and the second stage trains the Energy Matching process.
% (2) 给定一条采样出不符合约束的flow，我们可以认为，该flow和理想条件下的flow存在偏差，从而导致该flow末端点的样本不符合约束。 同时，该flow在采样过程中可以被视为一系列离散点的组合，假如直接纠正xt是否可以使其满足约束，相比于每一时间步进行约束，我们收到论文1的启发，通常在早期flow的不理想预测造成的影响更为严重，因此，我们直接纠正flow 过程的起点，图二显示了这种变化，即使是起点和终点sample差异巨大，但模型同样可以生成出合理轨迹。
% (3) 约束感知的训练：EBM  不同于diffusion-planner仅在训练时引入能量项，我们在训练阶段将约束编码为能量函数，当样本沿能量上升的方向采样时，样本则有更大的概率满足约束。具体而言，DAC约束，可有计算欧式符号距离场表达，越接近路沿，则该样本的能量越低。we 改进了 energy matching 进行模型训练。 一阶段训练 flow matching 过程，而第二阶段则训练 Energy matching 过程
\subsection{Reward as condition:}
% CATG尝试在推理过程中控制轨迹的激进程度。我们对navtrain 中的每一条gt轨迹进行navsim的仿真环境打分，打分获得的EP分数将作为condition控制轨迹激进程度。在推理过程中，我们直接将EP设置为一从而鼓励轨迹激进驾驶。
To control trajectory aggressiveness at inference time, CATG utilizes an EP (ego process) score as a conditioning signal. This score is derived by evaluating each GT trajectory in the NavTrain set within the NAVSIM simulator. By setting the EP condition to 1 during inference, the model is encouraged to produce more aggressive driving behavior.

\section{Experiments}
\subsection{Experiments Setup}
Our model is trained in two stages.
% 第一阶段训练包括训练flow matching过程以及感知模块以及地图分割模块。批量大小为64，学习率为2 10−4，共90个历元.
% 第二阶段训练我们遵循energy matching的第二阶段训练流程，仅仅训练flowmatching 过程，批量大小为64，学习率为2 10−4，共10个历元.在推理阶段，CATG推理出100条轨迹候选 with 100 sampling steps，并接入开源的预训练好的GTRS 打分器模型(V2-99 Backbone)进行打分。
The first stage of training encompasses the Flow Matching process, the perception module, and the map segmentation module. It was conducted with a batch size of 64, a learning rate of $2\times10^{-4}$, and trained for 90 epochs by using \textbf{NavTrain split}. The second stage of training adhered to the Energy Matching framework, focusing solely on fine-tuning the Flow Matching process. This stage used a batch size of 64, a learning rate of $2\times10^{-4}$, and trained for 10 epochs by using \textbf{NavTrain split}. During inference, CATG generates 100 candidate trajectories with 100 sampling steps.These candidates and trajectory vocabulary $vocab_{anchor}$ are then ranked by an open-source, pre-trained GTRS~\cite{GTRS} scorer model (with a V2-99 backbone) to select the most plausible trajectory as the final output.

\subsection{Experiments result}
We present our proposed CATG architecture’s results as shown in Tab.~\ref{tab:t1}.
\begin{table}[]
\caption{Results of proposed CATG architecture in NAVSIM V2}
\renewcommand\arraystretch{0.7}
\tabcolsep=0.8mm %%%%%%%%%
\resizebox{\linewidth}{!}{
\begin{tabular}{l|l}
\toprule
Metric Name                              & \multicolumn{1}{c}{\makecell{Team: bjtu\_jia\_team \\ \& qcraft}} \\
\midrule
extended pdm score combined              & \multicolumn{1}{c}{51.3116}      \\
\midrule
no at fault collisions stage one         & \multicolumn{1}{c}{98.2142}      \\
drivable area compliance stage one       & \multicolumn{1}{c}{100}      \\
driving direction compliance stage one   & \multicolumn{1}{c}{99.6428}      \\
traffic light compliance stage one       &        \multicolumn{1}{c}{100}                   \\
ego progress stage one                   &        \multicolumn{1}{c}{80.8379}                  \\
time to collision within bound stage one &        \multicolumn{1}{c}{98.5714}                   \\
lane keeping stage one                   &        \multicolumn{1}{c}{90}                   \\
history comfort stage one                &        \multicolumn{1}{c}{94.2857}                   \\
two frame extended comfort stage one     &        \multicolumn{1}{c}{57.1428}                   \\
no at fault collisions stage two         &        \multicolumn{1}{c}{88.9016}                   \\
drivable area compliance stage two       &        \multicolumn{1}{c}{95.4416}                   \\
driving direction compliance stage two   &        \multicolumn{1}{c}{97.9186}                   \\
traffic light compliance stage two       &        \multicolumn{1}{c}{96.8362}                   \\
ego progress stage two                   &        \multicolumn{1}{c}{77.9218}                   \\
time to collision within bound stage two &        \multicolumn{1}{c}{88.0227}                   \\
lane keeping stage two                   &        \multicolumn{1}{c}{56.6261}                   \\
history comfort stage two                &        \multicolumn{1}{c}{98.3082}                   \\
two frame extended comfort stage two     &    
\multicolumn{1}{c}{64.4264}                   \\
\bottomrule
\end{tabular}}
\label{tab:t1}
\end{table}

\section{Limitation}
Sampling trajectories with 100 steps remains computationally expensive. Nevertheless, accelerating this process may lead to a degradation in trajectory quality. Therefore, a promising direction for future work is to enhance sampling efficiency while preserving the quality of the generated trajectories.

\section{Conclusion}
We presents an end-to-end planner that leverages flow matching. Our approach is capable of incorporating flexible conditional signals to control trajectory generation. Furthermore, we innovatively propose three distinct strategies to enforce explicit constraints throughout the generation process. Experimental results presented in Tab.~\ref{tab:t1} demonstrate that our framework achieves a EPDMS of 51.31.

\small
\bibliographystyle{ieeenat_fullname}
\bibliography{main}
\end{document}